# A New Penta-valued Logic Based Knowledge Representation

**Vasile Patrascu**
Department of Informatics Technology, Tarom Company
Bucurestilor Road, 224F, Bucharest-Otopeni, Romania
e-mail: patrascu.v@gmail.com

**Abstract**

In this paper a knowledge representation model are proposed, FP5, which combine the ideas from fuzzy sets and penta-valued logic. FP5 represents imprecise properties whose accomplished degree is undefined, contradictory or indeterminate for some objects. Basic operations of conjunction, disjunction and negation are introduced. Relations to other representation models like fuzzy sets, intuitionistic, paraconsistent and bipolar fuzzy sets are discussed.

## 1   Introduction

Fuzzy sets are a specially well-suited tool to represent imprecise concepts with ill-defined boundaries. When a property $P$ is imprecise, its negation $\neg P$ is considered to be imprecise. The fuzzy set theory assumes both $P$ and $\neg P$ are related, namely: $\neg P(x) = 1 - P(x)$. However, this is not always true in real life. Hence, sometimes $P$ and $\neg P$ are represented independently. On the other hand, fuzzy sets do not allow to take into account the presence of objects whose membership degree $P$ is undefined. Three valued logics can solve the problem allowing three logical values: true, false and undefined. However, this is not always sufficient and we can find contradictions when a certain value $x$ verifies $P$ and $\neg P$ at the same time. Four-valued logics can solve the problem because it uses four logical values: true, false, undefined and contradictory. There is a special situation when the property $P$ and its negation $\neg P$ are close to 0.5 for some values of $x$. In this situation, we can detail the knowledge representation using a penta-valued logic based on five logical values: true, false, undefined, contradictory and indeterminate. In conclusion, the paper proposes a knowledge representation model, where fuzzy sets and penta-valued logic are combined to represent imprecise properties.

## 2   The Fuzzy Set and Its Extension

Let $X$ be a crisp set. In the framework of Zadeh theory [10], a *fuzzy set A* is defined by the membership function $\mu_A : X \to [0,1]$. The non-membership function $\nu_A : X \to [0,1]$ is obtained by negation and thus both functions define a partition of unity, namely:

$$\mu_A + \nu_A = 1 \qquad (2.1)$$

Atanassov has extended the fuzzy sets to the *intuitionistic fuzzy sets* [1]. Atanassov has relaxed the condition (2.1) to the following inequality:

$$\mu_A + \nu_A \leq 1 \qquad (2.2)$$

He has used the third function, the index of uncertainty $\pi_A$ that verifies the equality:

$$\pi_A = 1 - \mu_A - \nu_A \qquad (2.3)$$

Similarly, we can consider instead of (2.1) the following condition:

$$\mu_A + \nu_A \geq 1 \qquad (2.4)$$

Thus, we obtain the *paraconsistent fuzzy set* and one can define the index of contradiction :

$$\kappa_A = \mu_A + \nu_A - 1 \qquad (2.5)$$

There is a duality between intuitionistic fuzzy set and paraconsistent fuzzy sets. More generally, in this paper, we will consider as *bipolar fuzzy set* a set $A$, defined by two functions totally independent $\mu_A : X \to [0,1]$ and $\nu_A : X \to [0,1]$.

A *penta-valued fuzzy set* was defined based on Lukasiewicz penta-valued logic with the following five descriptors: strong membership, weak membership, index of uncertainty, weak non-membership and strong non-membership [9].

Belnap has defined a four-valued logic based on true, false, uncertainty and contradictory [2]. In this paper we will add to these four values the fifth: indeterminate. Thus, we will define a new penta-valued fuzzy set, constructing five-valued fuzzy partitions based on true, false, undefined, contradictory and indeterminate.

## 3 Transformation from Bipolar Knowledge Representation to a Penta-valued One

Let there be the Frank t-norm [5] defined for $s \in (0, \infty)$ by:

$$t_S(x, y) = \log_S\left(1 + \frac{(s^x - 1) \cdot (s^y - 1)}{s - 1}\right) \quad (3.1)$$

Let there be a Frank t-norm denoted by „∘". This t-norm verifies the Frank equation [5]:

$$x \circ y - \bar{x} \bullet \bar{y} = x + y - 1 \quad (3.2)$$

where $\bar{x}$ is the negation of $x$, namely:

$$\bar{x} = 1 - x$$

An equivalent form of Frank equation one can obtain by replacing $y$ with $\bar{y}$, namely:

$$x \circ \bar{y} - y \circ \bar{x} = x - y \quad (3.3)$$

Also, one defines its dual or its t-conorm „⊕" by:

$$x \oplus y = 1 - \bar{x} \circ \bar{y}$$

and thus, the formula (3.2) has the equivalent form:

$$x \oplus y + x \circ y = x + y$$

Let there be two t-norms „∘" and „•". We say that these two t-norms are *conjugated* if for $\alpha + \beta = 1$ there exists the equality:

$$x = x \bullet \alpha + x \circ \beta \quad (3.4)$$

Immediately, one results:

$$x \bullet y = x - x \circ \bar{y} \quad (3.5)$$

$$x \circ y = x - x \bullet \bar{y} \quad (3.6)$$

From (3.1) and (3.5) one obtains for the conjugate:

$$x \bullet y = \log_{\frac{1}{s}}\left(1 + \frac{\left(\frac{1}{s^x} - 1\right) \cdot \left(\frac{1}{s^y} - 1\right)}{\frac{1}{s} - 1}\right)$$

Thus, two Frank t-norms are conjugated if one is computed with parameter $s$ and the other is computed with parameter $\frac{1}{s}$. Thus, it results that the logics Godel and Lukasiewicz [6], [7], [8] are conjugated and the Product logic is identical with its conjugate.

For $s = 0$ it results:

$$\begin{cases} x \oplus y = Max(x, y) \\ x \circ y = Min(x, y) \\ x \bullet y = Max(0, x + y - 1) \end{cases} \quad (3.7)$$

From (3.5) one obtains

$$x = x \bullet \bar{y} + x \circ y \quad (3.8)$$

and replacing $x$ by $\bar{x}$ and $y$ by $\bar{y}$ it results:

$$\bar{x} = \bar{x} \bullet y + \bar{x} \circ \bar{y} \quad (3.9)$$

Replacing $x$ by $\bar{x} \circ \bar{y}$ and $y$ by $x \circ y$ in (3.8) it results:

$$\bar{x} \circ \bar{y} = \bar{x} \circ \bar{y} \bullet \overline{x \circ y} + \bar{x} \circ \bar{y} \circ x \circ y \quad (3.10)$$

or

$$\bar{x} \circ \bar{y} = (\bar{x} \circ \bar{y}) \bullet (\bar{x} \oplus \bar{y}) + \bar{x} \circ \bar{y} \circ x \circ y \quad (3.11)$$

Replacing $x$ by $x \circ y$ and $y$ by $\bar{x} \circ \bar{y}$ in (3.8) it results:

$$x \circ y = x \circ y \bullet \overline{\bar{x} \circ \bar{y}} + x \circ y \circ \bar{x} \circ \bar{y} \quad (3.12)$$

or

$$x \circ y = (x \circ y) \bullet (x \oplus y) + \bar{x} \circ \bar{y} \circ x \circ y \quad (3.13)$$

from (3.8) and (3.13) we obtain
$$x = x \bullet \bar{y} + (x \circ y) \bullet (x \oplus y) + \bar{x} \circ \bar{y} \circ x \circ y \quad (3.14)$$

and replacing $x$ by $y$ in (3.14) it results:

$$y = y \bullet \bar{x} + (x \circ y) \bullet (x \oplus y) + \bar{x} \circ \bar{y} \circ x \circ y \quad (3.15)$$

Now, we will denote:

$$\begin{cases} \tau = x \bullet \bar{y} \\ \varphi = \bar{x} \bullet y \\ \pi = (\bar{x} \circ \bar{y}) \bullet (\bar{x} \oplus \bar{y}) \\ \kappa = (x \circ y) \bullet (x \oplus y) \\ \iota = 2 \cdot (\bar{x} \circ \bar{y} \circ x \circ y) \end{cases} \quad (3.16)$$

From (3.8), (3.9), (3.11) and (3.13) it results a penta-valued partition of unity, namely:

$$\tau + \varphi + \kappa + \pi + \iota = 1 \quad (3.17)$$

From (3.14),(3.15) and (3.16) it results the inverse transform:

$$\begin{cases} x = \tau + \kappa + \dfrac{\iota}{2} \\ y = \varphi + \kappa + \dfrac{\iota}{2} \end{cases} \quad (3.18)$$

The functions defined by (3.16) have the following properties:

$$\begin{cases} \tau(x, y) = \varphi(y, x) \\ \tau(\bar{x}, \bar{y}) = \varphi(x, y) \end{cases} \quad (3.19)$$

$$\begin{cases} \iota(x, y) = \iota(y, x) \\ \iota(x, y) = \iota(\bar{x}, \bar{y}) \end{cases} \quad (3.20)$$

$$\begin{cases} \pi(x, y) = \pi(y, x) \\ \kappa(x, y) = \kappa(y, x) \\ \pi(\bar{x}, \bar{y}) = \kappa(x, y) \end{cases} \quad (3.21)$$

In the case of the pair of logics Lukasiewicz-Godel (3.7), one obtains the following particular forms for the parameters considered in (3.16).

$$\begin{cases} \tau = (x - y)_+ \\ \varphi = (y - x)_+ \\ \pi = (1 - x - y)_+ \\ \kappa = (x + y - 1)_+ \\ \iota = 1 - |x - y| - |x + y - 1| \end{cases} \quad (3.22)$$

where $a_+$ represents the positive part of $a$, namely:

$$a_+ = \frac{a + |a|}{2}$$

In bipolar preference theory $x$ could be the agreement function $S^+$ and $y$ could be the non-agreement function $S^-$ [3], [4]. Using (3.16) we can obtain $S^T, S^F, S^U, S^C, S^I$ which characterize five logical values, namely: true, false, undefined, contradictory and indeterminate. Thus, we came to have a penta-valued knowledge representation of bipolar imprecise information.

In the bipolar fuzzy set theory, $x$ could be the membership function $\mu_A$ and $y$ could be the non-membership function $\nu_A$. Using (3.16) we can obtain $\tau_A, \varphi_A, \kappa_A, \pi_A, \iota_A$ that define a penta-valued fuzzy set based on the five logical values that were mentioned above.

## 4 Penta-valued logic based on contradiction, undefinedness and indeterminacy

In the framework of this logic we will consider the following five logical: *true t*, *false f*, *undefined u*, *contradictory c* and *indeterminate i*. Tables 1, 2 and 3 show tables for basic operators in this logic.

Table 1: The **OR** operator

| ∨ | t | i | u | c | f |
|---|---|---|---|---|---|
| t | t | t | t | t | t |
| i | t | i | i | i | i |
| u | t | i | u | i | i |
| c | t | i | i | c | i |
| f | t | i | i | i | f |

Table 2: The **AND** operator

| ∧ | t | i | u | c | f |
|---|---|---|---|---|---|
| t | t | i | i | i | f |
| i | i | i | i | i | f |
| u | i | i | u | i | f |
| c | i | i | i | c | f |
| f | f | f | f | f | f |

One can see that the indeterminacy $i$ is absorbent for the uncertainty $u$ and contradiction $c$.

Table 3: The **NOT** operator

|   | ¬ |
|---|---|
| t | f |
| i | i |
| u | u |
| c | c |
| f | t |

The logical values *i, c, u* are the same with their negations.

## 5    Penta-valued Fuzzy Set

Let *X* be a crisp set. A penta-valued fuzzy set *A* is defined by the following functions: the membership $\tau_A : X \to [0,1]$, the non-membership $\varphi_A : X \to [0,1]$, the contradiction $\kappa_A : X \to [0,1]$ and the uncertainty $\pi_A : X \to [0,1]$. These functions verify the following inequality:

$$\tau_A + \varphi_A + \kappa_A + \pi_A \leq 1 \quad (5.1)$$

We will define the index of indeterminacy $\iota_A : X \to [0,1]$ by:

$$\iota_A = 1 - \tau_A - \varphi_A - \kappa_A - \pi_A \quad (5.2)$$

In this paper we will denote with FP5 the penta-valued fuzzy sets defined in this section.

For this kind of sets, one defines the union, the intersection and the negation operators.

### *The Union*

The union $A \cup B$ for two sets $A, B \in FP5$ is defined by formulae:

$$\begin{cases} \mu_{A \cup B} = \mu_A \vee \mu_B \\ \nu_{A \cup B} = \nu_A \wedge \nu_B \\ \pi_{A \cup B} = \pi_A \wedge \pi_B \\ \kappa_{A \cup B} = \kappa_A \wedge \kappa_B \end{cases} \quad (5.3)$$

### *The Intersection*

The intersection $A \cap B$ between two sets $A, B \in FP5$ is defined by the formulae:

$$\begin{cases} \mu_{A \cap B} = \mu_A \wedge \mu_B \\ \nu_{A \cap B} = \nu_A \vee \nu_B \\ \pi_{A \cap B} = \pi_A \wedge \pi_B \\ \kappa_{A \cap B} = \kappa_A \wedge \kappa_B \end{cases} \quad (5.4)$$

In formulae (5.3) and (5.4), the symbols "$\vee$" and "$\wedge$" represent any couple of t-conorm, t-norm.

### *The Complement*

The complement $A^c$ for the set $A \in FP5$ is defined by the formulae:

$$\begin{cases} \mu_{A^c} = \nu_A \\ \nu_{A^c} = \mu_A \\ \kappa_{A^c} = \kappa_A \\ \pi_{A^c} = \pi_A \end{cases} \quad (5.5)$$

In the set $\{0,1\}^4$ there are four vectors having the form $x = (\tau, \varphi, \kappa, \pi)$, which verify the condition (5.1): $T = (1,0,0,0)$ (*True*), $F = (0,1,0,0)$ (*False*), $C = (0,0,1,0)$ (*Contradictory*), $U = (0,0,0,1)$ (*Undefined*) and $I = (0,0,0,0)$ (*Indeterminate*).

Using the operators defined by (5.3), (5.4) and (5.5) the same table results as seen in Tables 1, 2 and 3.

## 6    Fuzzy Set and Its Extension as FP5

Imprecise concepts represented by bipolar fuzzy set can be translated into FP5. Particular forms of bipolar fuzzy sets can also be translated to FP5, for example fuzzy sets, intuitionistic fuzzy sets and paraconsistent fuzzy sets. The operations defined by (5.3), (5.4) and (5.5) supply new algebraic structures for the set types mentioned above.

### 6.1    Bipolar Fuzzy Set as FP5

One considers the bipolar fuzzy set $A \in BFS$ defined by the membership function $\mu_A$ and the non-membership function $\nu_A$. Using the formulae (3.22) one obtains:

$$\begin{cases} \tau_A = (\mu_A - \nu_A)_+ \\ \varphi_A = (\nu_A - \mu_A)_+ \\ \kappa_A = (\mu_A + \nu_A - 1)_+ \\ \pi_A = (1 - \mu_A - \nu_A)_+ \\ \iota_A = 1 - |\mu_A - \nu_A| - |\mu_A + \nu_A - 1| \end{cases} \quad (6.1.1)$$

The functions defined by (6.1.1) verify the condition (3.17) of a partition of unity.

We must underline that the index of indeterminacy $\iota_A$ is a symmetrical function. It has the maximum value in the point $(0.5, 0.5)$. This point is the center of the square defined by the points: $(0,0)$, $(1,0)$, $(0,1)$ and $(1,1)$. Also there are the following two equalities:

$$\begin{cases} \mu_A - \nu_A = \tau_A - \varphi_A \\ \mu_A + \nu_A = 1 + \kappa_A - \pi_A \end{cases} \quad (6.1.2)$$

From (3.18) it results the inverse transform:

$$\begin{cases} \mu_A = \tau_A + \kappa_A + \dfrac{\iota_A}{2} \\ \nu_A = \varphi_A + \kappa_A + \dfrac{\iota_A}{2} \end{cases} \quad (6.1.3)$$

Thus, we have transformed a bipolar knowledge representation into a penta-valued one.

### 6.2 Fuzzy Set as FP5

We consider the fuzzy set $A \in FS$ defined by the membership function $\mu_A$. One defines the non-membership function $\nu_A = 1 - \mu_A$. Using formulae (6.1.1) one define the indexes of truth, falsity and indeterminacy.

$$\begin{cases} \tau_A = (\mu_A - \nu_A)_+ \\ \varphi_A = (\nu_A - \mu_A)_+ \\ \iota_A = 1 - |\mu_A - \nu_A| \end{cases} \quad (6.2.1)$$

having the following equivalent forms:

$$\begin{cases} \tau_A = (2 \cdot \mu_A - 1)_+ \\ \varphi_A = (1 - 2 \cdot \mu_A)_+ \\ \iota_A = 1 - |2 \cdot \mu_A - 1| \end{cases} \quad (6.2.2)$$

Finally, due to the particularity (2.1) of fuzzy sets, the penta-valued representation is reduced to a three-valued one:

$$\tau_A + \varphi_A + \iota_A = 1$$

We must emphasize that the index of indeterminacy $\iota_A$ is a symmetrical function. It has the maximum value in the point $(0.5, 0.5)$. This point is the middle of the line between the points $(1,0)$ and $(0,1)$.

From (6.1.2) it results:

$$\begin{cases} \mu_A - \nu_A = \tau_A - \varphi_A \\ \mu_A + \nu_A = 1 \end{cases} \quad (6.2.3)$$

From (6.1.3) one obtains the inverse transform:

$$\begin{cases} \mu_A = \tau_A + \dfrac{\iota_A}{2} \\ \nu_A = \varphi_A + \dfrac{\iota_A}{2} \end{cases} \quad (6.2.4)$$

### 6.3 Intuitionistic Fuzzy Set as FP5

We consider the intuitionistic fuzzy set $A \in IFS$ defined by the membership function $\mu_A$ and the non-membership function $\nu_A$. We will translate to a penta-valued fuzzy set using formulae (6.1.1).

Thus, one defines the indexes of truth, falsity, uncertainty and indeterminacy.

$$\begin{cases} \tau_A = (\mu_A - \nu_A)_+ \\ \varphi_A = (\nu_A - \mu_A)_+ \\ \pi_A = 1 - \mu_A - \nu_A \\ \iota_A = \mu_A + \nu_A - |\mu_A - \nu_A| \end{cases} \quad (6.3.1)$$

Finally, due to the particularity (2.2) of intuitionistic fuzzy sets, the penta-valued representation is reduced to a tetra-valued one:

$$\tau_A + \varphi_A + \pi_A + \iota_A = 1$$

From (6.1.2) it results:

$$\begin{cases} \mu_A - \nu_A = \tau_A - \varphi_A \\ \mu_A + \nu_A = 1 - \pi_A \end{cases} \quad (6.3.2)$$

From (6.1.3) one obtains the inverse transform:

$$\begin{cases} \mu_A = \tau_A + \dfrac{\iota_A}{2} \\ \nu_A = \varphi_A + \dfrac{\iota_A}{2} \end{cases} \quad (6.3.3)$$

21

### 6.4 Paraconsistent Fuzzy Set as FP5

We consider the paraconsistent fuzzy set $A \in PFS$ defined by the membership function $\mu_A$ and the non-membership function $\nu_A$. We will translate to a penta-valued fuzzy set using formulae (6.1.1).

Thus, one defines the indexes of truth, falsity, contradiction and indeterminacy.

$$\begin{cases} \tau_A = (\mu_A - \nu_A)_+ \\ \varphi_A = (\nu_A - \mu_A)_+ \\ \kappa_A = \mu_A + \nu_A - 1 \\ \iota_A = 2 - |\mu_A - \nu_A| - \mu_A - \nu_A \end{cases} \quad (6.4.1)$$

Therefore, a bivalent knowledge representation was transformed into a tetravalent one, due to the particularity (2.4) of paraconsistent fuzzy sets.

The four defined indexes verify the partition of unity condition:

$$\tau_A + \varphi_A + \kappa_A + \iota_A = 1$$

From (6.1.2) one obtains:

$$\begin{cases} \mu_A - \nu_A = \tau_A - \varphi_A \\ \mu_A + \nu_A = 1 + \kappa_A \end{cases} \quad (6.4.2)$$

From (6.1.3) it results the inverse transform:

$$\begin{cases} \mu_A = \tau_A + \kappa_A + \dfrac{\iota_A}{2} \\ \nu_A = \varphi_A + \kappa_A + \dfrac{\iota_A}{2} \end{cases} \quad (6.4.3)$$

### 6 Conclusions

In this paper, a method was presented regarding multi-valued knowledge representation. The presented method is based on some properties of the Frank t-norms. Also, a new penta-valued logic was presented based on five logical values: true, false, undefined, contradictory and indeterminate. Using this new logic, new representations were obtained for fuzzy set, intuitionistic fuzzy set, paraconsistent fuzzy set and bipolar fuzzy set. These new representations and the operators defined on *FP5* supply new algebraic structures for these fuzzy sets types.